\documentclass[letterpaper, 10 pt, conference]{ieeeconf}  
\IEEEoverridecommandlockouts                              

\usepackage[utf8]{inputenc} 
\usepackage[T1]{fontenc}    
\usepackage{hyperref}       
\usepackage{url}            
\usepackage{booktabs}       
\usepackage{amsfonts}       
\usepackage{nicefrac}       
\usepackage{microtype}      
\usepackage{xcolor}         
\usepackage{graphicx}

\usepackage{microtype}
\usepackage{amssymb}
\usepackage{subcaption}

\usepackage{siunitx}   
\usepackage{makecell}
\usepackage{wrapfig}
\usepackage{tabularx}
\usepackage{amsmath}
\usepackage{soul}

\usepackage{listings}

\usepackage{xcolor}         

\newcommand{\changed}[1]{\textcolor{black}{#1}}

\title{\LARGE \bf
Scalable Policy Evaluation with Video World Models
}

\author{Wei-Cheng Tseng$^{1,2,3}$, Jinwei Gu$^{1}$, Qinsheng Zhang$^{1}$, Hanzi Mao$^{1}$, Ming-Yu Liu$^{1}$ \\
Florian Shkurti$^{2,3}$, Yen-Chen Lin$^{1}$
\thanks{$^{1}$Nvidia Research, $^{2}$University of Toronto, $^{3}$Vector Institute}%
}

\begin{document}

\maketitle
\thispagestyle{empty}
\pagestyle{empty}

\begin{abstract}
Training generalist policies for robotic manipulation has shown great promise, as they enable language-conditioned, multi-task behaviors across diverse scenarios. However, evaluating these policies remains difficult because real-world testing is expensive, time-consuming, and labor-intensive. It also requires frequent environment resets and carries safety risks when deploying unproven policies on physical robots. Manually creating and populating simulation environments with assets for robotic manipulation has not addressed these issues, primarily due to the significant engineering effort required and the substantial sim-to-real gap, both in terms of physics and rendering. In this paper, we explore the use of action-conditional video generation models as a scalable way to learn world models for policy evaluation. We demonstrate how to incorporate action conditioning into existing pre-trained video generation models. This allows leveraging internet-scale in-the-wild online videos during the pre-training stage and alleviates the need for a large dataset of paired video-action data, which is expensive to collect for robotic manipulation. Our paper examines the effect of dataset diversity, pre-trained weights, and common failure cases for the proposed evaluation pipeline. Our experiments demonstrate that, across various metrics, including policy ranking and the correlation between actual policy values and predicted policy values, these models offer a promising approach for evaluating policies without requiring real-world interactions.
\end{abstract}

\section{Introduction}
\label{introduction}
Robot learning has seen remarkable strides in the development of generalist robot manipulation policies that are capable of understanding natural language instructions and performing diverse tasks across different environments and robot embodiments~\cite{black2024pi_0,brohan2022rt,octo_2023,kim24openvla}. However, as these policies and the tasks they support grow in diversity and complexity, a fundamental challenge arises: how can we systematically evaluate the capabilities of generalist robot manipulation policies in a scalable and efficient way?

Current evaluation methodologies face significant obstacles that hinder rapid development and scalable deployment. Real-world testing and policy rollouts are the gold standard for estimating the performance of a policy, but they often incur high data collection and annotation costs, extensive robotic infrastructure requirements, limited reproducibility~\cite{kress2024robot}, and are severely limited by the need for environment resets. 
On the other hand, simulation-based approaches for policy evaluation require substantial manual effort to mitigate the physics and rendering sim-to-reality gap for robotic manipulation~\cite{funk2021benchmarking,li24simpler}. Constructing realistic simulations demands careful 3D modeling~\cite{li2024robogsimreal2sim2realroboticgaussian}, reward engineering~\cite{ma2023eureka}, and domain randomization~\cite{tobin2017domain}, yet these efforts often fail to bridge the sim-to-real gap in terms of physics accuracy and visual fidelity.

\begin{figure}[ht]
    \centering
    \includegraphics[width=1\linewidth]{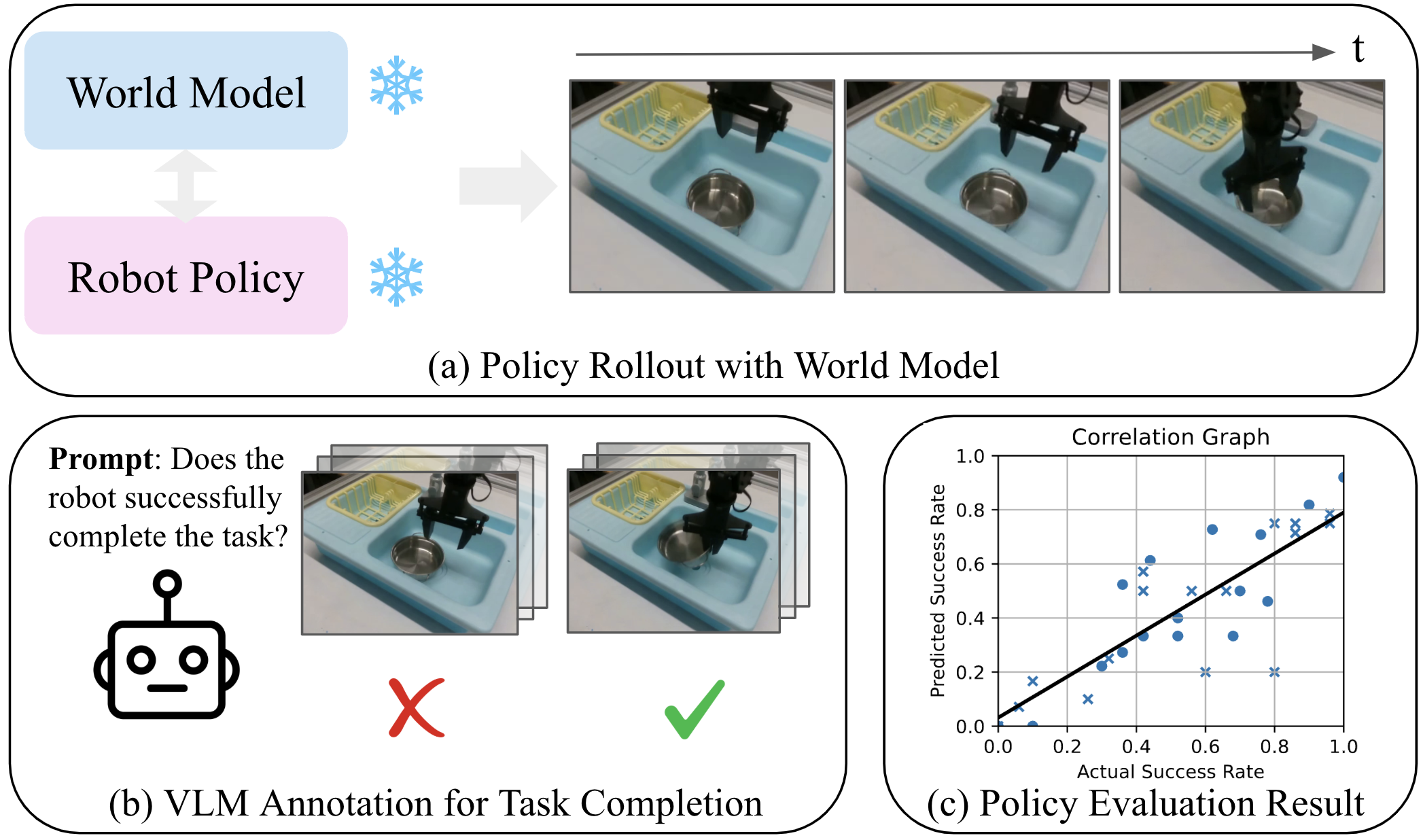}
    \caption{
    \textbf{Overview}.
    We propose using action-conditional video prediction models as world models for scalable policy evaluation. The process involves: (a) deploying a policy into the video model to generate rollout videos conditioned on 
    \changed{
    action sequences and initial frame, 
    }
    (b) using a VLM to judge task success or failure, and (c) computing the correlation between predicted and actual policy performance to evaluate the reliability of the approach.}
    \label{fig:teaser}
    \vspace{-10pt}
\end{figure}

Recently, action-conditional video prediction models~\cite{yang2023learning,alonso2024diffusionworldmodelingvisual,valevski2024diffusionmodelsrealtimegame,che2024gamegen,oasis2024,avid} have emerged as a promising tool for robotics to turn video prediction models into world models that can be used for robotics. These models learn to predict future frames given current observations and the agent's actions, providing a data-driven simulation framework for evaluating robot policies.
Compared to existing methods, this approach provides several key advantages: (1) unlike real-world testing, this method ensures reproducibility and does not require expensive physical equipment setups; (2) unlike handcrafted simulations, these models learn directly from real-world data, capturing complex visual and dynamic properties; (3) instead of manually constructing virtual environments to align the appearance of the environment and objects, these models condition directly on real observations, enabling effective real-to-sim transfer; (4) model performance improves with larger datasets and models, without requiring extensive human effort to engineer new tasks.

To this end, we propose a scalable, fully automated framework for policy evaluation that leverages video models. Our framework enables the policy to interact with an action-conditioned video prediction model~\cite{arxiv25_cosmos,nvidia_cosmos_predict2,wan2025,FangqiIRASim2024} to generate rollout videos. We then use an off-the-shelf Vision-Language Model (VLM)~\cite{gemini25_arxiv_2025,chatgpt2025} to annotate these rollouts and assess task completion, eliminating the need for human intervention and assessment. We present the correlation between real policy performance and estimated performance from the action-conditional video model, using both simulation and real-world environments and tasks.
Also, we systematically investigate the design space of diffusion-based action-conditional video prediction models and their effectiveness in evaluating robot policies. 
First, we examine how expanding the demonstration dataset (from 0.3 million to 1.1 million transitions) affects performance. 
While larger datasets improve generalization in terms of action-following, models still fail to enforce basic physical constraints, such as object permanence and rigid-body interactions.
Second, we study how pre-training on large-scale text-to-video and image-to-video datasets affects generalization. Our results demonstrate that pre-trained models yield more physically plausible predictions, underscoring the promise of large-scale pre-training for building data-driven simulators.
Third, we present experimental results using different pre-trained video model backbones. 
Finally, we highlight common failure cases in video models that may affect the reliability of policy evaluation.
Our contributions can be summarized as follows:
\begin{itemize}
\item We systematically evaluate action-conditional video prediction models for robot policy assessment, identifying their strengths and key failure modes, particularly in out-of-distribution action handling and physical realism.
\item
\changed{
We conduct an analysis of the effects of dataset diversity and pre-training of action-conditional video prediction models for policy evaluation tasks.
}
\item
\changed{
We plan to release the proposed policy evaluation pipeline, including the initial frames for both real-world and synthetic tasks, the action-conditioned model, and the VLM for task completion, to facilitate further benchmarking of action-conditioned video models in policy evaluation.
}
\end{itemize}
Supplementary materials, including additional qualitative results and an appendix, are available on \href{https://miscsubmission.github.io/world_model_policy_eval/}{the project website}.

\section{Related Work}
\textbf{Video Generation Models.} 
Recent advancements in diffusion and autoregressive models have significantly improved video generation capabilities~\cite{openai2024video,lin2024open,yang2024cogvideox,polyak2024movie,arxiv25_cosmos,wan2025}. 
These models can generate videos conditioned on diverse input modalities such as text, images, depth maps, segmentation masks, camera poses, and motion trajectories~\cite{geng2024motionprompting,he2024cameractrl}.
Building on top of this progress, a growing body of work explores how video models conditioned on actions can function as interactive simulators, supporting applications in entertainment~\cite{che2024gamegen,bruce2024genie,gamengen}, planning~\cite{du2023video}, and reinforcement learning~\cite{yang2023learning,alonso2024diffusionworldmodelingvisual}. This progress has spurred interest in their potential integration into embodied AI systems. In this work, we systematically investigate the impact of scaling model size, dataset diversity, and pre-training tasks on the effectiveness of using action-conditioned video models as a data-driven simulator for robot policy evaluation.

\textbf{Robot Policy Evaluation.} 
Reliable robot policy evaluation is essential for benchmarking algorithmic progress and ensuring transparency in real-world performance. In autonomous driving, policies are commonly validated in simulation across diverse scenarios before real-world deployment~\cite{dosovitskiy2017carla,yang2023unisim}. However, simulating dynamic interactions with objects, realistic materials, and complex lighting conditions in robotic manipulation remains an open challenge.
Efforts such as YCB~\cite{calli2015ycb} have standardized objects for evaluation, yet standardizing the remaining factors within an environment remains a challenge. 
Real-world robot challenges, including the Amazon Picking Challenge~\cite{correll2016analysis} and the DARPA Robotics Challenge~\cite{krotkov2018darpa}, address this by using fixed physical test sites. 
However, these setups demand extensive preparation and maintenance. These challenges motivate increasing interest in using simulation to estimate policies' success rates.
SIMPLER~\cite{li24simpler} constructs physical simulators to approximate real-world environments. They show that one can deploy the policies trained with real-world data in the simulator to estimate their performance in the real world. 
However, this approach has two major limitations: (1) the reliance on extensive domain randomization to close the sim-to-real gap, and (2) the need for manually curated 3D data and precise environment design. In contrast, our approach eliminates manual engineering by leveraging action-conditional video prediction models to generate simulation environments directly from policies' observations, offering a scalable and data-driven alternative to classical simulator-based evaluation methods.
AutoEval~\cite{arxiv25_autoeval} presents a system that autonomously evaluates generalist robot policies in the real world. However, real-world policy evaluation still faces several challenges, such as motor overheating and long wall-clock times. Moreover, real-world evaluations are inherently imperfect and subject to noise and variability.
WorldEval~\cite{worldeval} proposed using a video model for policy evaluation; however, it requires the video model to utilize features from the action decoder as conditioning signals, which limits its applicability for evaluating new policies.

\section{Method}
\subsection{Evaluating Robot Policies}
Let $N$ be the total number of policies. Let $\pi_i$ and $\pi_j$ be a pair of them, with real-world cumulative performance $R_i$ and $R_j$, for example, their average success rates across a representative set of tasks, as estimated by real-world rollouts. Our goal is to construct a world simulator $S$, for which there is a strong correlation between the relative performances $R_i$ and $R_j$ in the real world and the relative performances in simulation $R_{S,i}$ and $R_{S,j}$.

\begin{figure*}[t]
    \centering
    \includegraphics[width=1\linewidth]{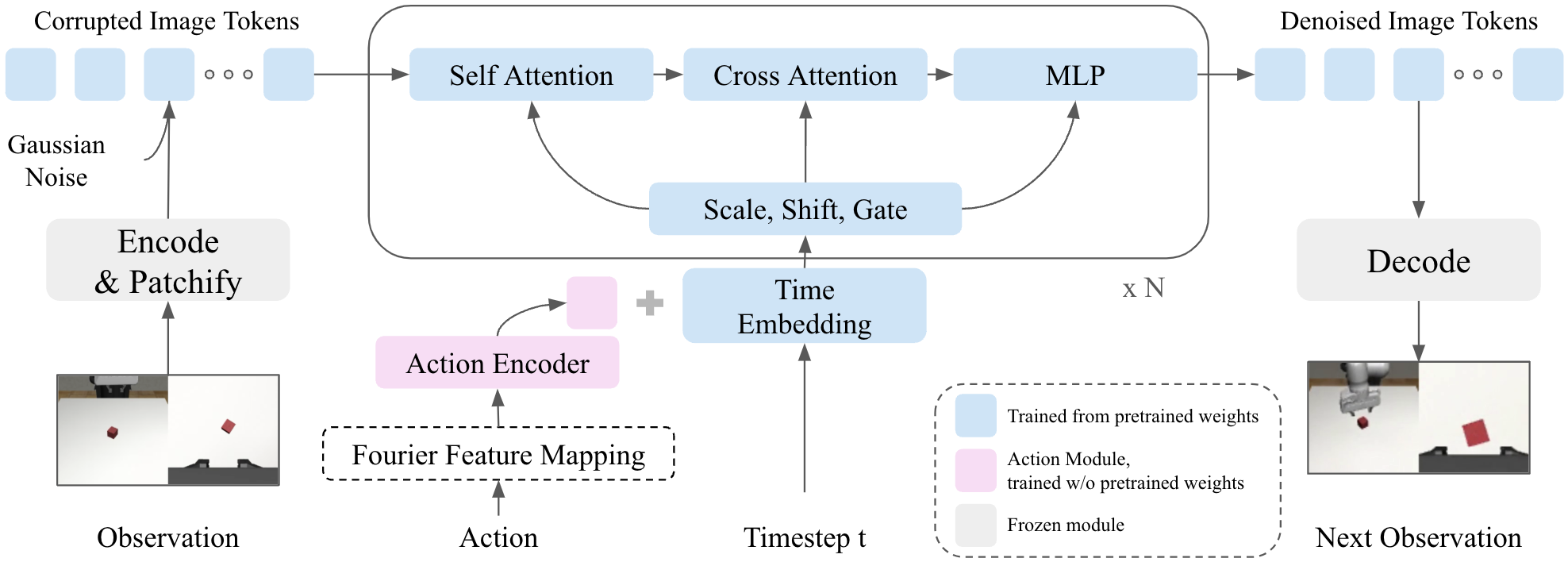}
    \vspace{-14pt}
    \caption{\textbf{Network Architecture for Action-Conditional Video Prediction with a DiT-Based Video Model such as Cosmos-Predict2-2B~\cite{nvidia_cosmos_predict2}.} We process the raw action information with Fourier feature mappings~\cite{fourier} and encode it into a latent action token. The latent action token is added to a time-embedding, which enables us to make the video diffusion model action-conditional.
    }
    \label{fig:main_network_arch}
    \vspace{-16pt}
\end{figure*}
\emph{Metrics.} To evaluate our approach, we use the metrics proposed by SIMPLER~\cite{li24simpler}, including MMRV and the Pearson correlation coefficient~\cite{pearson1895vii}. Specifically, MMRV (Mean Maximum Rank Violation) is defined as follows:
\begin{align}
\text{RankViolation}(i, j) 
&= |R_i - R_j| \cdot \mathbf{1}\!\Big[ \\
&\quad (R_{S, i} < R_{S, j}) \neq (R_i < R_j) \Big] \nonumber
\end{align}
\begin{equation}
    \text{MMRV}(R, R_S) = \frac{1}{N} \sum_{i=1}^N \max_{1 \leq j \leq N} \text{RankViolation}(i, j)
\end{equation}

\noindent The key concept underlying MMRV is \textit{rank violation} between two policies, $\pi_i$ and $\pi_j$, which quantifies the extent to which the world model misranks the policies relative to their true performance margins in the real world. MMRV aggregates the $N^2$ rank violations by computing the average worst-case rank violation for each policy.


Once we have a world model $S$, we can roll out the policy autoregressively by allowing it to interact with the world model. 
After the completion of a model-based rollout, we leverage Vision-Language Models~\cite{iclr25_in_context_value,colm24_task_success} to assess whether the policy successfully completes the task. To be more specific, we subsample frames from the rollout video and feed them into VLM, a prompt is also provided to identify the definition of task completion. For the detailed prompting design of the VLM, we refer the reader to the supplemental material on the project website.

\subsection{Learning Action-conditional Video Prediction Models}
Let $\mathcal{O}$ be the space of observations (e.g., video frames), and let $\mathcal{A}$ be the space of actions, which is the end-effector pose of the robot arm in world coordinates. At each time step $t$, a policy $\pi$ receives a sequence of past observations $o_{1:t} = \{o_1, ..., o_t\} \in \mathcal{O}$ and decides on an action $a_t \in A$. This chosen action is then executed in the environment, leading to the observation of the next step $o_{t+1} \in O$.

Our goal is to build an action-conditional video prediction model that forecasts $o_{t+1}$ given $o_{1:t}$ and $a_{1:t}$. Formally, we aim to learn the \textit{dynamics} of the environment $p(o_{t+1} \mid o_{1:t}, a_{1:t})$.
This approach is task-agnostic, as multiple tasks often share the same observation space (video frames). Consequently, the learned model can leverage data from multiple tasks, potentially improving performance through knowledge transfer. Moreover, it provides a consistent simulator that can be applied broadly across tasks.
By iteratively sampling from this predictive model---each time conditioning on its own previously generated observations---
we can simulate a complete interaction trajectories. 

\emph{Video Model Architectures and Implementation.} 
We leverage the open-sourced pre-trained video diffusion model, Cosmos-Predict2-Video2World-2B~\cite{arxiv25_cosmos} (hereafter referred to as Cosmos), as the backbone of our action-conditional video prediction models.
Cosmos-Predict2-Video2World-2B is a Diffusion Transformer~\cite{peebles2023scalable}, but it has additional cross-attention layers between self-attention and feedforward layers for conditioning. It is pre-trained with text-to-video and image-to-video on large-scale dataset. In total, it has 2B trainable parameters. 
Cosmos-series models, like many other video generation models, are not designed by default to
condition on actions. On one hand, this has the benefit of allowing learning from in-the-wild video
data online; on the other hand, this is problematic for robotics simulation, which requires conditioning on actions.
To augment the model with action-conditioning capability, we add action embeddings to the time embeddings. The action embeddings are generated by first 
\changed{
applying Fourier feature mappings~\cite{fourier} to the raw action signals to enable the network to learn high-frequency functions from low-dimensional inputs}
 %
%
, followed by processing them through a MLP.
The FLOPs of this MLP are significantly smaller than the rest of the network, so the modification doesn’t affect the inference and training efficiency. The model architecture is summarized in Figure \ref{fig:main_network_arch}, with additional details in appendix VII
.
The entire framework is trained end-to-end using the EDM objective~\cite{Karras2022edm,karras2024analyzingimprovingtrainingdynamics}, following the same setup as Cosmos~\cite{arxiv25_cosmos,nvidia_cosmos_predict2}.

\begin{table*}[t]
    \caption{\textbf{Quantitative Results for Policy Evaluation in Synthetic Setting}. Pearson correlation coefficient (higher is better) and MMRV (Mean Metric Rank Violation; lower is better) between video model and simulator policy evaluations.
    }
    \vspace{-4pt}
    \centering
    \small 
    \renewcommand{\arraystretch}{0.6}
    \setlength{\tabcolsep}{0pt} 
    \begin{tabularx}{\textwidth}{l *{8}{>{\centering\arraybackslash}X}}
        \toprule
        \textbf{World Model} & \multicolumn{2}{c}{\textbf{Lift}} & \multicolumn{2}{c}{\textbf{ Can}} & \multicolumn{2}{c}{\textbf{Square}} & \multicolumn{2}{c}{\textbf{Tool Hang}} \\
        \cmidrule(lr){2-3} \cmidrule(lr){4-5} \cmidrule(lr){6-7} \cmidrule(lr){8-9}
        & {Pearson} $\uparrow$ & {MMRV} $\downarrow$
        & {Pearson} $\uparrow$ & {MMRV} $\downarrow$
        & {Pearson} $\uparrow$ & {MMRV} $\downarrow$
        & {Pearson} $\uparrow$ & {MMRV} $\downarrow$ \\
        \midrule
        \makecell[l]{\texttt{Cosmos-Predict2-2B (Ours)}} 
            & \textbf{0.879} & \textbf{0.015} & \textbf{0.860} & \textbf{0.141} & \textbf{0.847} & \textbf{0.165} & \textbf{0.833} & \textbf{0.217} \\
        \midrule
        \makecell[l]{{\quad w/o Policy Rollouts}} 
            & 0.856 & 0.222 & 0.744 & 0.226 & 0.505 & 0.384 & 0.502 & 0.288 \\
        \makecell[l]{{\quad w/o Pre-trained Weights}} 
            & 0.822 & 0.220 & 0.716 & 0.233 & 0.523 & 0.245 & 0.645 & 0.363 \\
        \midrule
        \makecell[l]{\texttt{Cosmos-Predict-1-7B}} 
            & 0.845 & 0.188 & 0.750 & 0.231 & 0.606 & 0.204 & 0.500 & 0.281 \\
        \makecell[l]{\texttt{Cosmos-Predict-1-14B}} 
            & 0.800 & 0.260 & 0.716 & 0.233 & 0.511 & 0.261 & 0.666 & 0.360 \\
        \makecell[l]{\texttt{Wan-2.1-14B-LoRA}} 
            & 0.842 & 0.180 & 0.746 & 0.166 & 0.801 & 0.179 & 0.789 & 0.244 \\
        \midrule
        \texttt{IRASim} & 0.835 & 0.103 & 0.780 & 0.230 & 0.213 & 0.708 & 0.642 & 0.330 \\
        \bottomrule
    \end{tabularx}
    \label{tab:policy_eval_quantity_all}
\end{table*}
\begin{figure*}[t]
    \centering
    \vspace{-4pt}
    \includegraphics[width=1\linewidth]{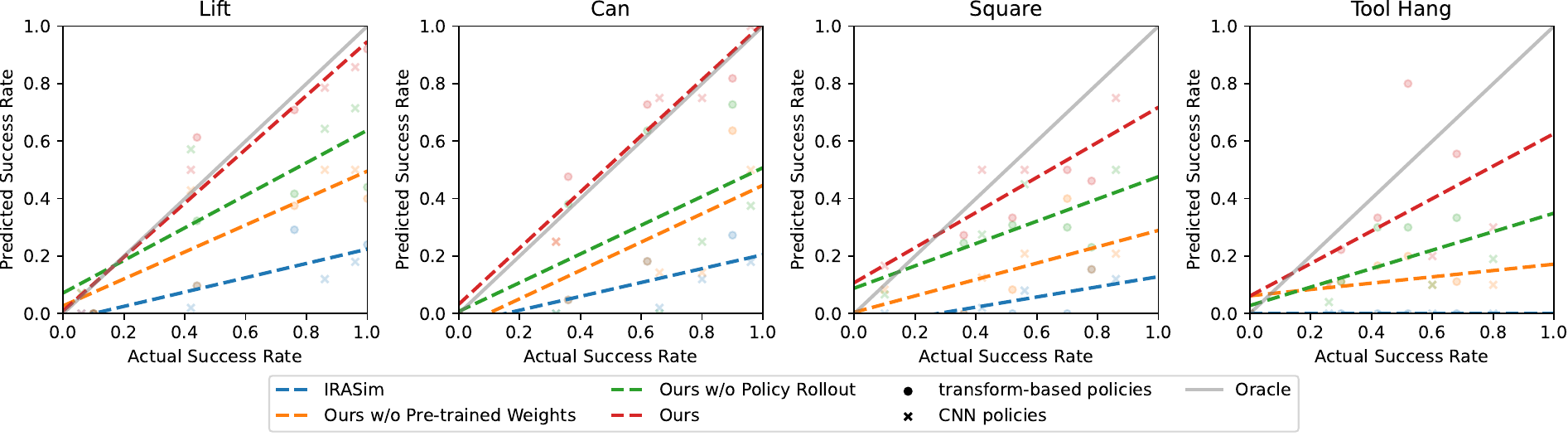}
    \caption{\textbf{Correlation Plots for Policy Evaluation}. 
    We plot the correlation between predicted policy performance via evaluation in the video model vs. actual policy performance for the simulation tasks.}
    \label{fig:correlation_graph_model_scale}
    \label{fig:corr_prior}
    \label{fig:corr_dataset_scale}
    \vspace{-8pt}
\end{figure*}

\subsection{Policy Rollout Data Augmentation}
To train the video model we use demonstration trajectories collected by human teleoperation. However, the action distribution within the demonstration trajectories could be limited, for instance, the trajectories in the demonstration dataset typically correspond to successful trajectories, and failures are rare.
To address this, we augment the demonstration dataset with rollouts from policies with different manipulation performance to enhance the generalization capability of the action-conditional video prediction model.

\section{Experiments} \label{sec:experiments}

\subsection{Synthetic Setting}
\begin{figure}[t]
    \centering
    \includegraphics[width=1\linewidth]{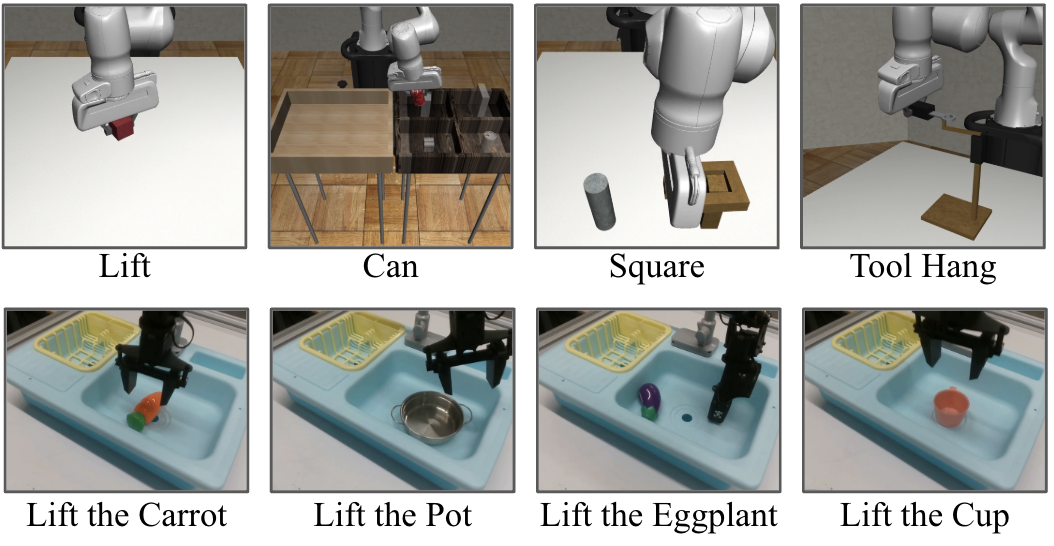}
    \vspace{-12pt}
    \caption{\textbf{Manipulation Tasks}. We evaluate the proposed approach in both synthetic and real-world tasks. In the synthetic setting, we mainly consider four tasks in RoboMimic~\cite{corl21_robomimic}: lift, can, square and tool hang. For real-world experiments, we consider four tasks from the Bridge benchmark: lifting a carrot, a pot, an eggplant, and a cup. 
    }
    \label{fig:tasks}
    \vspace{-6pt}
\end{figure}
\noindent 
\changed{For a controllable study}, we first test our policy evaluation paradigm in a synthetic setting\changed{, where the environment dynamics are treated as ground truth}. In this case, the dynamics in the synthetic environment are treated as ground-truth dynamics. Specifically, we ask whether the video model can predict the performance ranking of policies trained entirely on synthetic data and evaluated in the same synthetic environment.

We consider the following four tasks in RoboMimic~\cite{corl21_robomimic}:
\begin{itemize} 
\item \textbf{Lift}: The robot lifts a small cube. This is the simplest task.
\item \textbf{Can}: The robot moves a Coke can from a large bin to a smaller target bin. It is harder than Lift because grasping the can is less stable, and precise placement is required.
\item \textbf{Square}: 
The robot places a square nut onto a rod. This is more challenging than Lift and Can, as it requires precise grasping and careful alignment.
\item \textbf{Tool Hang}: The robot assembles a frame by inserting a hook into a base and then hanging a wrench on the hook. This is the most complex task, involving multiple stages of precise, dexterous, and rotation-heavy movements.
\end{itemize}

\begin{table*}[t]
    \centering
    \small 
    \caption{\textbf{Quantitative Results for Action-Conditional Video Prediction.} We generate 64 future frames in an auto-regressive manner and evaluate on both RoboMimic (in-distribution) and policy rollout (out-of-distribution) trajectories.
    }
    \vspace{-4pt}
    \setlength{\tabcolsep}{1pt} 
    \begin{tabular*}{\textwidth}{@{\extracolsep{\fill}}lcccccccc@{}}
        \toprule
        \textbf{World Model} & \multicolumn{4}{c}{\textbf{RoboMimic (In-Distribution)}} & \multicolumn{4}{c}{\textbf{Policy Rollouts (Out-of-Distribution)}} \\
        \cmidrule(lr){2-5} \cmidrule(lr){6-9}
        & PSNR $\uparrow$ & SSIM $\uparrow$ & Latent L2 $\downarrow$ & FVD $\downarrow$
        & PSNR $\uparrow$ & SSIM $\uparrow$ & Latent L2 $\downarrow$ & FVD $\downarrow$ \\
        \midrule
        \makecell[l]{\texttt{Cosmos-Predict2-2B }} & \textbf{20.622} & \textbf{0.868} & \textbf{0.315} & \textbf{98.455} & \textbf{15.996} & \textbf{0.793} & \textbf{0.458} & \textbf{250.234} \\
        \midrule
        \makecell[l]{{\quad w/o Policy Rollouts}} & 18.712 & 0.841 & 0.362 & 130.211 & 14.931 & 0.783 & 0.461 & 285.001 \\
        {\quad w/o Pre-trained Weights} & 17.731 & 0.834 & 0.384 & 127.695 & 14.754 & 0.766 & 0.519 & 291.158 \\
        \bottomrule
    \end{tabular*}
    \label{tab:robot_action_conditional_video_prediction_all}
    \vspace{-4pt}
\end{table*}

\begin{figure*}[t]
    \centering
    \begin{subfigure}[b]{0.435\textwidth}
        \centering
        \includegraphics[width=\textwidth]{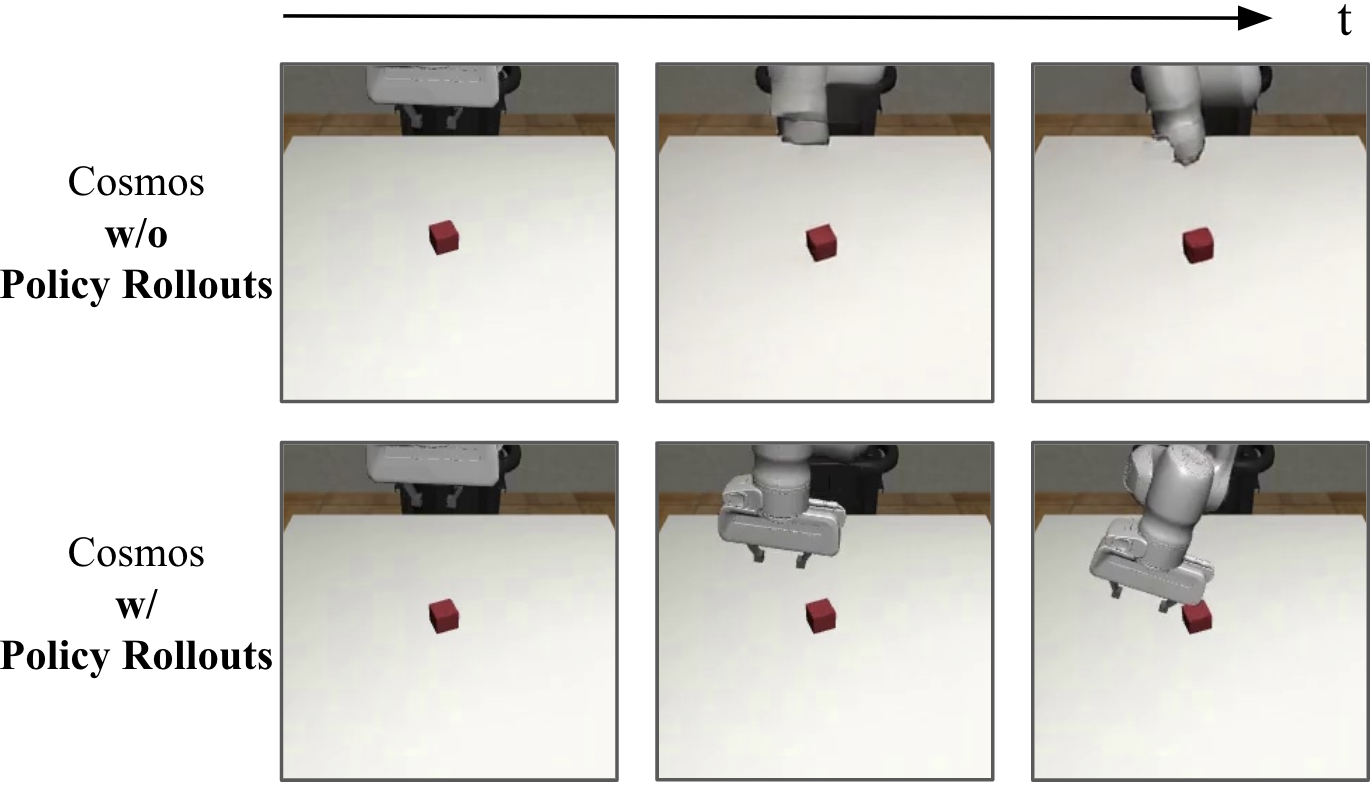}
        \caption{Comparison between with and without policy rollouts.}
        \label{fig:ood_video_prediction}
    \end{subfigure}
    \hfill
    \begin{subfigure}[b]{0.555\textwidth}
        \centering
        \includegraphics[width=\textwidth]{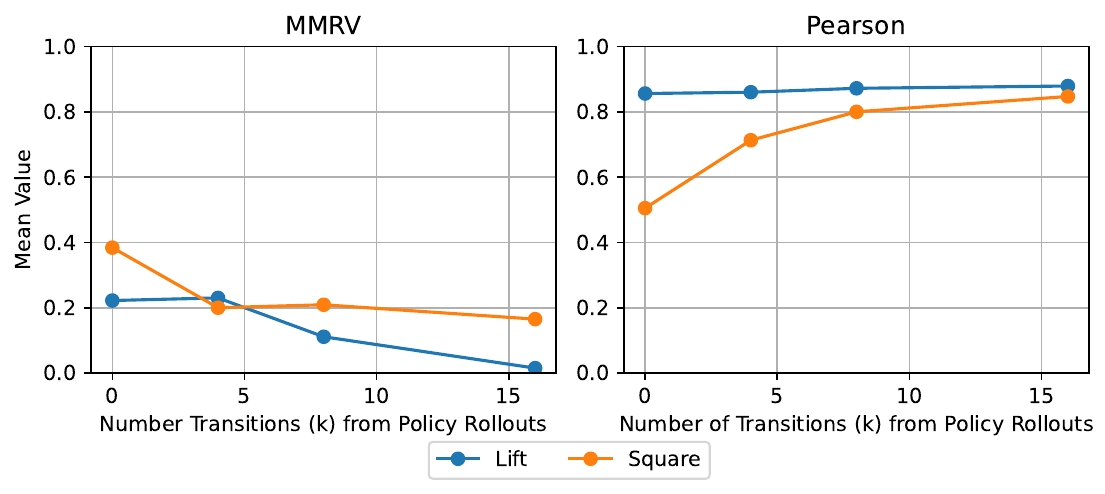}
        \caption{Effect of Policy Rollouts on Policy Evaluation}
        \label{fig:rollout_scaling}
    \end{subfigure}
    \caption{\textbf{Effect of Policy Rollouts on Policy Evaluation}. (a) Without policy rollouts, predictions degrade over time—for example, the gripper gradually disappears. Incorporating rollouts significantly improves visual quality and realism. (b)  Increasing the number of policy rollouts in the video model’s training set consistently enhances policy evaluation performance.}
    \vspace{-8pt}
\end{figure*}

\noindent We train action-conditional video prediction models using both human demonstration trajectories and policy rollouts for four tasks from the RoboMimic~\cite{corl21_robomimic} simulation environment. The original RoboMimic dataset contains 1,700 successful demonstrations. To expand this dataset, we add rollouts from transformer-based policies trained in Robosuite~\cite{robosuite2020}. These rollouts introduce more diverse behaviors than human demonstrations\footnote{Importantly, the policies used to generate rollouts for training are not reused in later policy evaluation experiments.}. A summary of the augmented dataset is provided in 
Appendix VIII
. 
The combined dataset is split into training, validation, and test sets.

For evaluation, we measure video prediction quality with PSNR~\cite{psnr}, SSIM~\cite{ssim}, latent L2, and FVD~\cite{iclr19_fvd}. Latent L2 and PSNR capture reconstruction error in latent and pixel space, respectively. SSIM measures similarity in brightness, contrast, and structure, while FVD compares feature distributions to assess overall video quality.

For policy evaluation, we train a set of policies using diffusion policy~\cite{rss23_diffusion_policy} with either CNN- or transformer-based architectures on RoboMimic~\cite{corl21_robomimic} data. Each policy takes two image inputs (a gripper view and an agent view) from the most recent two timesteps and predicts the next eight actions. To create policies of varying quality, we train them for different numbers of steps.

Table~\ref{tab:policy_eval_quantity_all} presents the quantitative results of evaluating these policies with the Cosmos model, and Figure~\ref{fig:corr_prior} shows the corresponding correlation plots. Overall, our approach achieves sufficient accuracy to provide reliable policy ranking signals. That said, correlation is generally weaker for long-horizon tasks (e.g., Tool Hang and Square) than for shorter ones (e.g., Lift and Can). Video prediction performance is reported separately in Table~\ref{tab:robot_action_conditional_video_prediction_all}.

\subsection{Ablation Study}

To better understand the factors that affect video models for policy evaluation, we study the role of training data, pre-trained weights, and different model initializations.

\subsubsection{Effectiveness of Data Augmentation}
We investigate whether adding policy rollouts to the training set improves the video diffusion model. For comparison, we also train a baseline model using only human demonstration trajectories.

As shown in Table~\ref{tab:robot_action_conditional_video_prediction_all}, adding policy rollouts—trajectories generated by executing policies in RoboMimic—improves video prediction performance, especially on out-of-distribution (OOD) data. Figure~\ref{fig:ood_video_prediction} further shows that models trained without rollouts produce more frequent artifacts. For policy evaluation, excluding rollouts leads to a sharp drop in the correlation between actual and predicted policy performance (Figure~\ref{fig:corr_prior}, Table~\ref{tab:policy_eval_quantity_all}), which we attribute to lower-quality video predictions when interacting with policies.

We also analyze how the number of policy rollouts affects evaluation on the Lift and Square tasks (Figure~\ref{fig:rollout_scaling}). In general, more rollouts improve evaluation accuracy, with the effect being more pronounced for the harder Square task than for the simpler Lift task.


\begin{figure*}[t]
    \centering
    \begin{subfigure}[b]{0.65\textwidth}
        \centering
        \includegraphics[width=\textwidth]{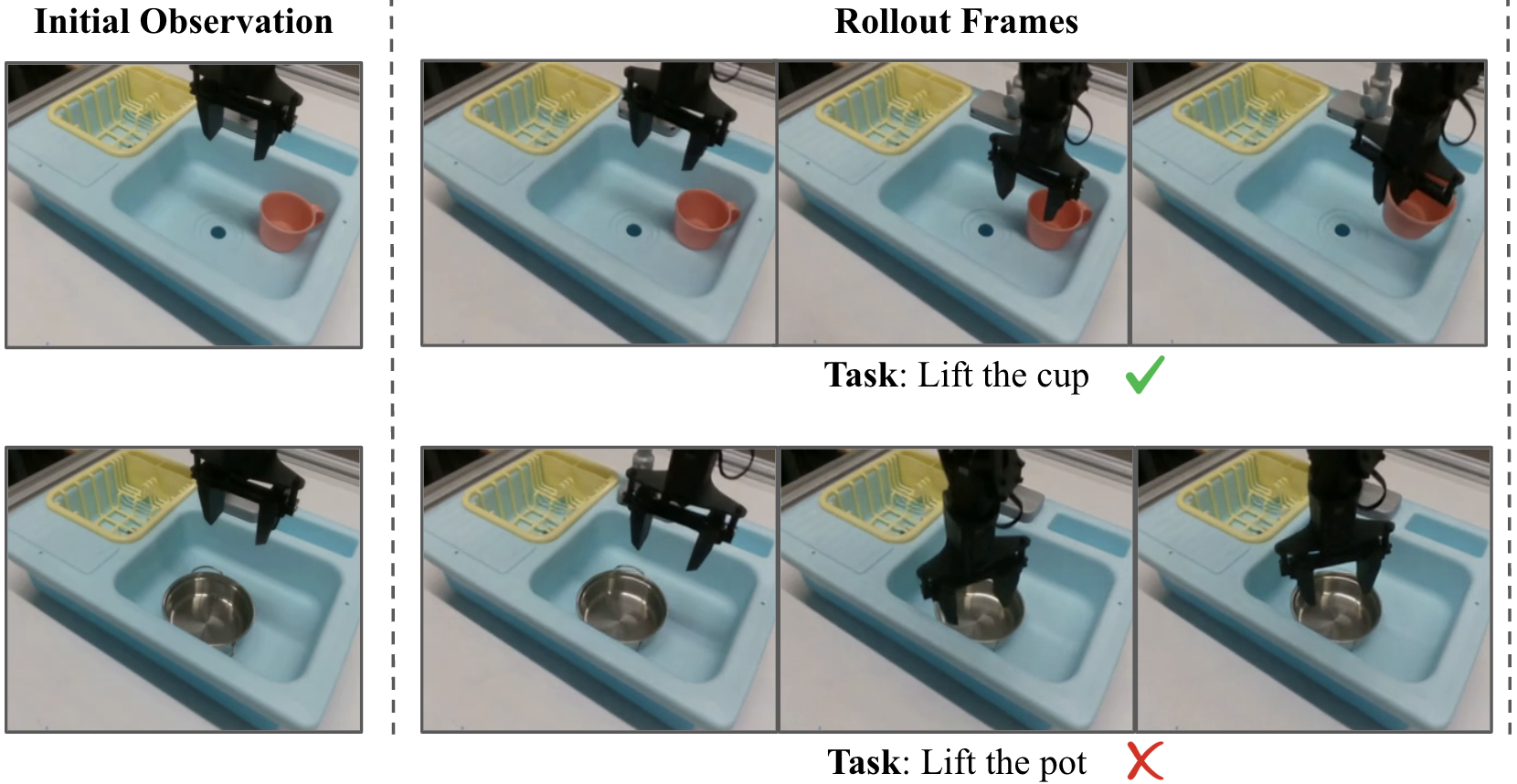}
        \caption{ Qualitative Result for Policy Rollout on Video Model}
    \end{subfigure}
    \hfill
    \begin{subfigure}[b]{0.34\textwidth}
        \centering
        \includegraphics[width=\textwidth]{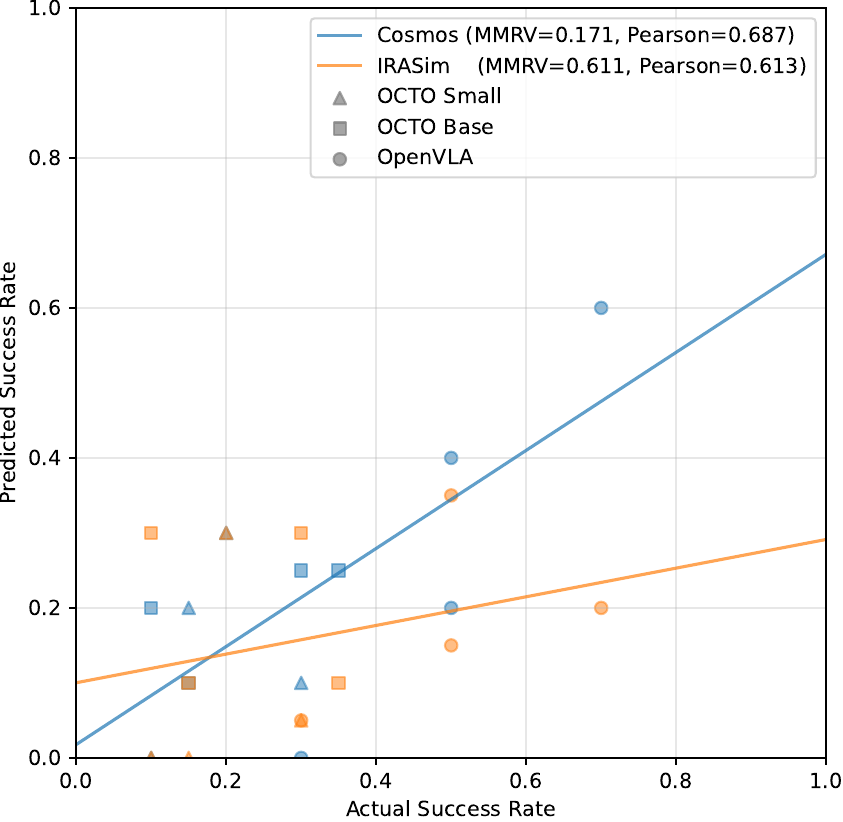}
        \caption{Correlation Graph}
    \end{subfigure}
    \caption{\textbf{Policy Evaluation on the Bridge Dataset~\cite{bridge_v2,bridge_v1}}
    (a) 
    We roll out the policy Octo-Small~\cite{octo_2023} with the action-conditional video model, trained on the Bridge dataset. 
    \changed{
    The top row shows a successful lift of the cup, while the bottom row illustrates a failure rollout where the policy is unable to lift the pot.
    }
    (b) 
    Our video model can serve as an evaluation environment, and the performance acquired from it has positive correlation with real-world performance.
    }
    \label{fig:bridge_quan}
    \vspace{-6pt}
\end{figure*}

\subsubsection{Effectiveness of Pretrained Weights}
To assess the impact of pretraining, we compare models trained from scratch against those initialized with the pre-trained weights of Cosmos-Predict2-Video2World-2B~\cite{arxiv25_cosmos}, which was trained on a large, high-quality text-to-video dataset for physical-AI applications.

As shown in Table~\ref{tab:robot_action_conditional_video_prediction_all}, removing pre-trained weights causes a substantial drop in video prediction quality for both in-distribution and out-of-distribution data. Models trained from scratch exhibit more frequent distortions and hallucinations compared to pre-trained models.

\begin{figure}
    \centering
    \includegraphics[width=1\linewidth]{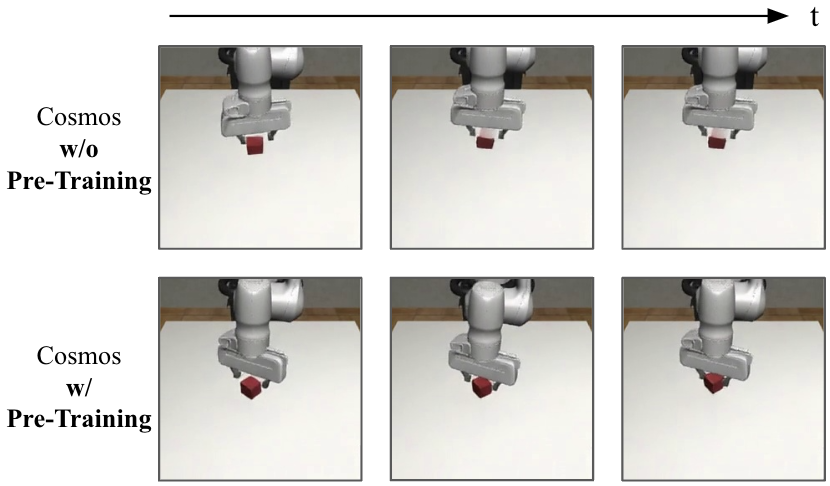}
    \caption{Pretrained weights mitigate object disappearance during interaction, unlike randomly initialized weights.}
    \label{fig:prior}
    \vspace{-12pt}
\end{figure}

We hypothesize that pretraining provides useful priors about physical dynamics that are difficult to learn from demonstration data alone. Without these priors, models trained from scratch often display hallucination, deformation, and occlusion when the gripper interacts with objects or when objects collide (see Figure~\ref{fig:prior}). This suggests that the model without pretraining lacks a robust understanding of physics, limiting its effectiveness for policy evaluation.

Policy evaluation results, summarized in Figure~\ref{fig:corr_prior} and Table~\ref{tab:policy_eval_quantity_all}, confirm that pre-trained priors significantly improve evaluation accuracy.

\subsubsection{Different Pre-trained Models}
We benchmark policy evaluation across several video model backbones, including Wan2.1~\cite{wan2025} fine-tuned with LoRA, Cosmos-Predict1~\cite{arxiv25_cosmos}, and IRASim~\cite{FangqiIRASim2024}. Among these, Cosmos-Predict2—pre-trained on physical-AI data—consistently achieves the best performance. Its advantage likely comes from the strong alignment between its pretraining dataset and the physical-AI domain, which enables better generalization to robotics tasks compared to Wan2.1, whose pretraining was on general video data.

In contrast, IRASim, which is trained entirely from scratch without pretraining, performs the worst. This further underscores the importance of pre-trained weights for effective policy evaluation. Implementation details for post-training across different models are provided in 
Appendix VI-A.2
.

\subsection{Policy Evaluation in Real-World Setup}

We next examine whether our policy evaluation approach generalizes to real-world environments and generalist policies. Following Simpler~\cite{li24simpler}, we evaluate OCTO-Small, OCTO-Base~\cite{octo_2023}, and OpenVLA~\cite{kim24openvla} in the Bridge setup~\cite{bridge_v2,bridge_v1}. The evaluated tasks include "lift the pot", "lift the carrot", "lift the eggplant", and "lift the cup".

For training the video model, we use the Bridge V2 dataset~\cite{bridge_v1,bridge_v2}. To better align the original Bridge setup with our replicated setup, we additionally collect 300 trajectories in our own environment and include them in the training dataset.

As shown in Figure~\ref{fig:bridge_quan}, our world model provides an effective evaluation environment: policy performance predicted by the model correlates positively with actual performance in the real world. Further details on the real-world experiments are provided in Appendix
VI-A
.

\subsection{Failure Case Analysis} \label{sec:limitation}

\begin{figure*}
    \centering
    \includegraphics[width=1\linewidth]{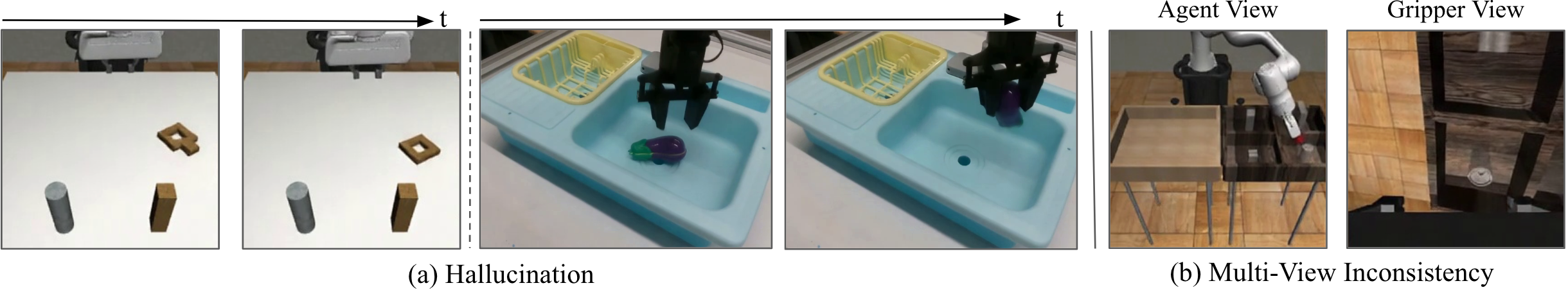}
    \vspace{-16pt}
    \caption{\textbf{Common Failure Cases in Policy Evaluation with Video Models} (a) \textbf{Hallucination}. The first instance shows that the handle of the square disappears due to a hallucination. This would lead to failure of policy. 
    \changed{
    The other instance shows that Cosmos will hallucinate the eggplant into the gripper, which could lead to overestimation of policy performance. 
    }
%
    (b) \textbf{Multi-View Inconsistency}. The generated result may fail to ensure multi-view consistency. For instance, in the agent view, the can appears to be grasped by the gripper, but it is missing in the gripper view. 
    }
    \label{fig:common_failure}
    \vspace{-8pt}
\end{figure*}
We have demonstrated the potential and feasibility of using video prediction models for policy evaluation. In this section, we discuss the limitations of our approach and highlight common failure modes observed in both simulation and real-world experiments. More quantitative analysis about failure cases and the capacity of the base model is provided in the appendix.

\emph{Hallucinations.} 
Even after post-training the video prediction model on our demonstration dataset to make it action-conditional and improve physical plausibility, we still observe problems such as inconsistent object behavior and lack of object permanence (Figure~\ref{fig:common_failure}(a)). These artifacts can introduce bias into policy evaluation. In addition, because rollout videos are generated in an autoregressive manner, errors accumulate over time. This compounding effect becomes especially severe in long-horizon tasks—for example, each rollout of the Tool Hang task involves 700 transitions, where small prediction errors can quickly escalate.

\emph{Multi-View Inconsistency.}
Our world model takes two concatenated inputs (the gripper view and the agent view) and predicts the corresponding two views at the next timestep. However, the predicted views can be inconsistent~\cite{asim24met3r}, with the same object appearing differently across viewpoints (Figure~\ref{fig:common_failure}(b)). This issue likely stems from the absence of an explicit view-consistency loss during training. As with hallucinations, the problem is amplified in the autoregressive setting, where small inconsistencies accumulate over time.

\emph{Replicated Prediction.}
When the gripper’s displacement is very small, the diffusion model often replicates the conditional image as the predicted image. This leads to a static environment, which hinders policy evaluation by preventing the policy from progressing toward achieving its task. The issue is more common in Tool Hang. Because insertion requires extremely precise, fine-grained motions, the gripper’s displacements are minimal, making replication more likely.

\emph{VLM Annotation Errors.}
Automating task completion judgments with VLM is imperfect. Often, VLM annotations achieve only 65–80\% accuracy. A more detailed analysis of these annotation errors is provided in Appendix
VI-A.5
.

\section{Conclusion}
\label{conclusion}

We investigated how to adapt pre-trained video prediction models into action-conditional world models for scalable robot policy evaluation. By conditioning large video models on actions, our approach enables policy assessment without the need for costly real-world rollouts or heavily engineered simulation environments. Our analysis underscores three key factors for reliable evaluation: the value of diverse training datasets, the benefits of pre-trained weights, and the failure modes that can arise during prediction.

Across both synthetic and real-world experiments, action-conditional video models accurately predict policy performance, achieving strong correlations with ground-truth success rates and supporting policy ranking. These results point to video-based world models as a promising foundation for data-driven evaluation pipelines, reducing reliance on expensive robotic infrastructure and accelerating progress toward generalist robotic policies.








\bibliography{example_paper}{}
\bibliographystyle{IEEEtran.bst}

\newpage
\newpage


\end{document}